\documentclass{llncs}

\usepackage[utf8]{inputenc}
\usepackage{amsmath}
\usepackage[authoryear,round]{natbib}
\usepackage{xcolor}
\usepackage{xspace}
\usepackage{listings}

\definecolor{rulecolor}{rgb}{0.80,0.80,0.80}

\definecolor{htmltitlecolor}{rgb}{0.23, 0.04, 0}
\definecolor{darkred}{rgb}{0.7, 0, 0}

\definecolor{SkoleBrown}{RGB}{150,100,50}
\definecolor{SkoleYellow}{RGB}{232,183,3}
\definecolor{SkoleOrange}{RGB}{212,80,0}
\colorlet{SkoleLightBrown}{SkoleBrown!50!SkoleYellow}
\definecolor{darkgreen}{cmyk}{0.7, 0, 1, 0.5}
\definecolor{lightgreen}{rgb}{.2, 1, .2}
\definecolor{darkblue}{cmyk}{1, 0.8, 0, 0}
\definecolor{lightblue}{rgb}{.4, .6, 1,}
\definecolor{grey}{cmyk}{0.1,0.1,0.1,1}
\definecolor{lightgrey}{cmyk}{0,0,0,0.5}
\definecolor{purple}{cmyk}{0.8,1,0,0}

\colorlet{titlecolor}{SkoleOrange!50!black}
\colorlet{codeblockcolorbase}{SkoleLightBrown!80!black}
\colorlet{codeblockcolor}{codeblockcolorbase!02}

\newcommand{\sklearn}{\textit{scikit-learn}\xspace}

\title{API design for machine learning software: experiences from the
scikit-learn project}

\author{Lars~Buitinck~\inst{1} \and
        Gilles~Louppe~\inst{2} \and
        Mathieu~Blondel~\inst{3} \and
        Fabian~Pedregosa~\inst{4} \and
        Andreas~C.~M\"uller~\inst{5} \and
        Olivier~Grisel~\inst{6} \and
        Vlad~Niculae~\inst{7} \and
        Peter~Prettenhofer~\inst{8} \and
        Alexandre~Gramfort~\inst{4,9} \and
        Jaques~Grobler~\inst{4} \and
        Robert~Layton~\inst{10} \and
        Jake~Vanderplas~\inst{11} \and
        Arnaud~Joly~\inst{2} \and
        Brian Holt~\inst{12} \and
        Gaël~Varoquaux~\inst{4}}

\institute{ILPS, Informatics Institute, University of Amsterdam \and
           University of Liège \and
           Kobe University \and
           Parietal, INRIA Saclay \and
           University of Bonn \and
           Independent consultant \and
           University of Bucharest \and
           Ciuvo GmbH \and
           Institut Mines-Telecom, Telecom ParisTech, CNRS LTCI \and
           University of Ballarat \and
           University of Washington \and
           Samsung Electronics Research Institute}

\DeclareRobustCommand{\VAN}[3]{#2}

\begin{document}

\maketitle

\begin{abstract}
\sklearn is an increasingly popular machine learning
library. Written in Python, it is designed to be simple and efficient, accessible to
non-experts, and reusable in various contexts. In this paper, we present and
discuss our design choices for the application programming interface (API) of
the project. In particular, we describe the simple and elegant interface shared
by all learning and processing units in the library and then discuss its
advantages in terms of composition and reusability. The paper also comments on
implementation details specific to the Python ecosystem and analyzes obstacles
faced by users and developers of the library.
\end{abstract}

\setcounter{footnote}{0}

\section{Introduction}

The \sklearn project\footnote{\url{http://scikit-learn.org}}
\citep{pedregosa2011} provides an open source machine learning
library for the Python programming language. The ambition of the project is to
provide efficient and well-established machine learning tools within a
programming environment that is accessible to non-machine learning experts and
reusable in various scientific areas. The project is not a novel domain-specific
language, but a library that provides machine learning idioms to a
general-purpose high-level language. Among other things, it includes classical learning
algorithms, model evaluation and selection tools, as well as
preprocessing procedures. The library is
distributed under the simplified BSD license, encouraging its use in both
academic and commercial settings.

\sklearn is a library, i.e.\ a collection of classes
and functions that users import into Python programs. Using \sklearn therefore
requires basic Python programming knowledge. No command-line interface, let
alone a graphical user interface, is offered for non-programmer users. Instead,
interactive use is made possible by the Python interactive interpreter, and its
enhanced replacement IPython \citep{perez2007ipython}, which offer a
\textsc{matlab}-like working environment specifically designed for scientific
use.

The library has been designed to tie in with the set of numeric
and scientific packages centered around the NumPy and SciPy libraries. NumPy
\citep{vanderwalt2011} augments Python with a contiguous numeric array datatype
and fast array computing primitives,
while SciPy \citep{varoquaux2013scipy} extends it
further with common numerical operations, either by implementing
these in Python/NumPy or by wrapping existing C/C{}\verb!++!/Fortran
implementations. Building upon this stack, a series of libraries called
\textit{scikits} were created, to complement SciPy with
domain-specific toolkits. Currently, the two most popular and feature-complete
ones are by far \sklearn and
\textit{scikit-image},\footnote{\url{http://scikit-image.org}}
which does image processing.

Started in 2007, \sklearn is developed by an international team of over a dozen
core developers, mostly researchers from various fields (e.g.,
computer science, neuroscience, astrophysics). The project also benefits
from many occasional contributors proposing small bugfixes or
improvements. Development proceeds on GitHub\footnote{\url{https://github.com/scikit-learn}},
a platform which greatly facilitates this kind of
collaboration. Because of the large number of developers, emphasis is
put on keeping the project maintainable. In particular, code must follow
specific quality guidelines, such as style consistency and unit-test coverage.
Documentation and examples are required for all features,
and major changes must pass code review by at least two
developers not involved in the implementation of the proposed change.

\sklearn's popularity can be gauged from various indicators such as the hundreds
of citations in scientific publications, successes in various machine learning
challenges (e.g., Kaggle), and statistics derived from our
repositories and mailing list.  At the time of writing, the project is watched
by 1365 people and forked 693 times on GitHub; the mailing list receives more
than 300 mails per month; version control logs
show 183 unique contributors to the codebase and the online documentation
receives 37,000 unique visitors and 295,000 pageviews per month.

\citet{pedregosa2011} briefly presented \sklearn and
benchmarked it against several competitors.
In this paper, we instead present an
in-depth analysis of design choices made when building the library,
detailing how we organized and operationalized
common machine learning concepts.
We first present in section~\ref{sec:core-api} the central application
programming interface (API) and then describe, in section~\ref{sec:advanced-api},
advanced API mechanisms built on the core interface.
Section~\ref{sec:implementation} briefly describes the implementation.
Section~\ref{sec:comparison} then
compares \sklearn to other major projects in terms of API\@.
Section~\ref{sec:future_work} outlines some of the objectives for
a \sklearn 1.0 release.
We conclude by summarizing the major points of this paper in
section~\ref{sec:conclusions}.

\section{Core API}

\label{sec:core-api}

All objects within \sklearn share a uniform common basic API consisting of three
complementary interfaces: an \textit{estimator} interface for building and
fitting models, a \textit{predictor} interface for making predictions and a
\textit{transformer} interface for converting data. In this section, we describe
these three interfaces, after reviewing our general principles and data
representation choices.

\subsection{General principles}

As much as possible, our
design choices have been guided so as to avoid the proliferation of framework
code. We try to adopt simple conventions and to limit to a minimum the number of
methods an object must implement. The API is designed to adhere to the following
broad principles:

\begin{description}
  \item[Consistency.]
       All objects (basic or composite) share a consistent interface composed of
       a limited set of methods. This interface is documented in a consistent
       manner for all objects.
  \item[Inspection.]
       Constructor parameters and parameter values determined by learning
       algorithms are stored and exposed as public attributes.
  \item[Non-proliferation of classes.]
       Learning algorithms are the only objects to be represented using custom classes.
       Datasets are represented as NumPy arrays or SciPy sparse matrices.
       Hyper-parameter names and values are represented as standard
       Python strings or numbers whenever possible.
       This keeps \sklearn easy to use and easy to combine with other libraries.
  \item[Composition.]
       Many machine learning tasks are expressible
       as sequences or combinations of transformations to data.
       Some learning algorithms are also naturally viewed
       as meta-algorithms parametrized on other algorithms.
       Whenever feasible, such algorithms are implemented and composed from
       existing building blocks.
  \item[Sensible defaults.]
       Whenever an operation requires a user-defined parameter,
       an appropriate default value is defined by the library.
       The default value should cause the operation to be performed
       in a sensible way (giving a baseline solution for the task at hand).
\end{description}

\subsection{Data representation}
\label{sec:arrays}

In most machine learning tasks, data is modeled as a set of variables.  For
example, in a supervised learning task, the goal is to find a mapping
from input variables $X_1, \ldots X_p$, called features, to
some output variables $Y$. A sample is then defined as a pair of
values $([x_1, \ldots, x_p]^\mathrm{T}, y)$ of these variables. A widely used
representation of a dataset, a collection of
such samples, is a pair of matrices with numerical values: one for the input
values and one for the output values. Each row of these matrices
corresponds to one sample of the dataset and each column to one variable of
the problem.

In \sklearn, we chose a representation of data that is as close as
possible to the matrix representation: datasets are encoded as NumPy
multidimensional arrays for dense data and as SciPy sparse matrices for sparse
data. While these may seem rather unsophisticated data representations when
compared to more object-oriented constructs, such as the ones used by
Weka \citep{hall2009weka}, they bring the prime advantage of allowing us to rely
on efficient NumPy and SciPy vectorized operations while keeping
the code short and readable.  This design choice has also been motivated by
the fact that, given their pervasiveness in many other scientific Python
packages, many scientific users of Python are already familiar with NumPy dense
arrays and SciPy sparse matrices.
From a practical point of view, these formats also provide a collection of
data loading and conversion tools which make them very easy to use in many
contexts. Moreover, for tasks where the inputs are text files or semi-structured
objects, we provide \textit{vectorizer} objects that efficiently convert such
data to the NumPy or SciPy formats.

For efficiency reasons, the public interface is oriented towards processing
batches of samples rather than single samples per API call. While classification
and regression algorithms can indeed make predictions for single samples,
\sklearn objects are not optimized for this use case. (The few online learning
algorithms implemented are intended to take mini-batches.) Batch processing makes
optimal use of NumPy and SciPy by preventing the overhead inherent to Python
function calls or due to per-element dynamic type checking. Although this might
seem to be an artifact of the Python language, and therefore an implementation
detail that leaks into the API, we argue that APIs should be designed so as not
to tie a library to a suboptimal implementation strategy. As such, batch
processing enables fast implementations in lower-level languages (where memory
hierarchy effects and the possibility of internal parallelization come into
play).

\subsection{Estimators}
\label{sec:estimators}

The \textit{estimator} interface is at the core of the
library. It defines instantiation mechanisms of objects and exposes a
\texttt{fit} method for learning a model from training data.  All supervised and
unsupervised learning algorithms (e.g., for classification, regression or
clustering) are offered as objects implementing this interface. Machine
learning tasks like feature extraction, feature selection or dimensionality
reduction are also provided as estimators.

Estimator initialization and actual learning are strictly separated,
in a way that is similar to partial function application:
an estimator is initialized from a set of named constant hyper-parameter values
(e.g., the $C$ constant in SVMs)
and can be considered as a function
that maps these values to actual learning algorithms.
The constructor of an estimator does not see any actual data, nor does it perform any actual learning.
All it does is attach the given parameters to the object.
For the sake of convenient model inspection, hyper-parameters are set as public attributes,
which is especially important in model selection settings.
For ease of use, default hyper-parameter values are also provided
for all built-in estimators.
These default values are set to be relevant in many common
situations in order to make estimators as effective as possible
\textit{out-of-box} for non-experts.

Actual learning is performed by the \texttt{fit} method. This method is called
with training data (e.g., supplied as two arrays \texttt{X\_train} and
\texttt{y\_train} in supervised learning estimators). Its task is to run a
learning algorithm and to determine model-specific parameters from the training
data and set these as attributes on the estimator object. As a convention, the
parameters learned by an estimator are exposed as public attributes with names
suffixed with a trailing underscore (e.g., \texttt{coef\_} for the
learned coefficients of a linear model),
again to facilitate model inspection.
In the partial application view,
\texttt{fit} is a function from data to a model of that data.
It always returns the estimator object it was called on,
which now serves as a model of its input and can be used to perform predictions or transformations of input data.

From the start, the choice to let a single object serve dual purpose as
estimator and model has mostly been driven by usability and technical
considerations. From the user point of view, having two coupled instances (i.e.,
an estimator object, used as a factory, and a model object, produced by the
estimator) indeed decreases the ease of use and is also more likely to
unnecessarily confuse newcomers. From the developer point of view, decoupling
estimators from models also creates parallel class hierarchies and increases the
overall maintenance complexity of the project. For these practical reasons, we
believe that decoupling estimators from models is not worth the effort. A good
reason for decoupling however, would be that it makes it possible to ship a
model in a new environment without having to deal with potentially complex
software dependencies. Such a feature could however still be implemented in
\sklearn by making estimators able to export a fitted model, using the
information from its public attributes, to an agnostic model description such as
PMML~\citep{guazzelli2009pmml}.

To illustrate the initialize-fit sequence,
let us consider a supervised learning task using logistic regression.
Given the API defined above, solving this problem is as simple as the following
example.
\begin{lstlisting}
from sklearn.linear_model import LogisticRegression

clf = LogisticRegression(penalty="l1")
clf.fit(X_train, y_train)
\end{lstlisting}
In this snippet, a \texttt{LogisticRegression} estimator is first initialized by
setting the \texttt{penalty} hyper-parameter to \texttt{"l1"} for
$\ell_1$ regularization. Other hyper-parameters (such as \texttt{C},
the strength of the regularization) are not explicitly given and
thus set to the default values. Upon calling \texttt{fit}, a model is
learned from the training arrays \texttt{X\_train} and \texttt{y\_train},
and stored within the object for later use.
Since all estimators share the same interface, using a different learning algorithm is
as simple as replacing the constructor (the class name);
to build a random forest on
the same data, one would simply replace
\texttt{LogisticRegression(penalty="l1")} in the snippet above by
\texttt{RandomForestClassifier()}.

In \sklearn, classical learning algorithms are not the only objects to be
implemented as estimators. For example, preprocessing routines (e.g., scaling of
features) or feature extraction techniques (e.g., vectorization of text
documents) also implement the \textit{estimator} interface. Even stateless
processing steps, that do not require the \texttt{fit} method to
perform useful work, implement the estimator interface. As we will illustrate
in the next sections, this design pattern is indeed of prime importance for
consistency, composition and model selection reasons.

\subsection{Predictors}

The \textit{predictor} interface extends the notion of an estimator
by adding a \texttt{predict}
method that takes an array \texttt{X\_test} and produces
predictions for \texttt{X\_test}, based on the learned parameters of the
estimator (we call the input to \texttt{predict} ``\texttt{X\_test}'' in order
to emphasize that \texttt{predict} generalizes to new data). In the case of
supervised learning estimators, this method typically returns the predicted
labels or values computed by the model.  Continuing with the previous example,
predicted labels for \texttt{X\_test} can be obtained using the following
snippet:
\begin{lstlisting}
y_pred = clf.predict(X_test)
\end{lstlisting}

Some unsupervised learning estimators may also implement the \texttt{predict}
interface. The code in the snippet below fits a $k$-means model with $k=10$ on
training data \texttt{X\_train}, and then uses the  \texttt{predict} method to
obtain cluster labels (integer indices) for unseen data \texttt{X\_test}.
\begin{lstlisting}
from sklearn.cluster import KMeans

km = KMeans(n_clusters=10)
km.fit(X_train)
clust_pred = km.predict(X_test)
\end{lstlisting}

Apart from \texttt{predict}, predictors may also implement methods
that quantify the confidence of predictions. In the case of
linear models, the \texttt{decision\_function} method returns
the distance of samples to the separating hyperplane. Some
predictors also provide a \texttt{predict\_proba} method which returns
class probabilities.

Finally, predictors must provide a \texttt{score} function to assess their
performance on a batch of input data. In supervised estimators, this method
takes as input arrays \texttt{X\_test} and \texttt{y\_test} and typically
computes the coefficient of determination between \texttt{y\_test} and
\texttt{predict(X\_test)} in regression, or the accuracy
in classification.
The only requirement is that the \texttt{score} method return a value
that quantifies the quality of its predictions (the higher, the better).
An unsupervised estimator may also expose a \texttt{score} function
to compute, for instance, the likelihood of the given data under its model.

\subsection{Transformers}

Since it is common to modify or filter data before feeding it to a learning
algorithm, some estimators in the library implement a \textit{transformer}
interface which defines a \texttt{transform} method. It takes as input some new
data \texttt{X\_test} and yields as output a transformed version of
\texttt{X\_test}. Preprocessing, feature selection, feature extraction and dimensionality reduction
algorithms are all provided as transformers within the library.  In our example,
to standardize the input \texttt{X\_train} to zero mean and unit variance
before fitting the logistic regression estimator,
one would write:
\begin{lstlisting}
from sklearn.preprocessing import StandardScaler

scaler = StandardScaler()
scaler.fit(X_train)
X_train = scaler.transform(X_train)
\end{lstlisting}
Of course, in practice, it is important to apply the same preprocessing to the
test data \texttt{X\_test}. Since a \texttt{StandardScaler} estimator stores the
mean and standard deviation that it computed for the training set, transforming
an unseen test set \texttt{X\_test} maps it into the appropriate region of
feature space:
\begin{lstlisting}
X_test = scaler.transform(X_test)
\end{lstlisting}
Transformers also include a variety of learning algorithms, such as
dimension reduction (PCA, manifold learning), kernel approximation,
and other mappings from one feature space to another.

Additionally, by leveraging the fact that \texttt{fit} always returns the
estimator it was called on, the \texttt{StandardScaler} example above can be
rewritten in a single line using method chaining:
\begin{lstlisting}
X_train = StandardScaler().fit(X_train).transform(X_train)
\end{lstlisting}

Furthermore, every transformer allows \texttt{fit(X\_train).transform(X\_train)}
to be written as \texttt{fit\_transform(X\_train)}.
The combined \texttt{fit\_transform} method prevents repeated computations.
Depending on the transformer,
it may skip only an input validation step,
or in fact use a more efficient algorithm for the transformation.
In the same spirit, clustering algorithms provide a
\texttt{fit\_predict} method
that is equivalent to \texttt{fit} followed by \texttt{predict},
returning cluster labels assigned to the training samples.

\section{Advanced API}

\label{sec:advanced-api}

Building on the core interface introduced in the previous section, we now
present advanced API mechanisms for building meta-estimators,
composing complex estimators and selecting models. We also discuss design
choices allowing for easy usage and extension of \sklearn.

\subsection{Meta-estimators}

Some machine learning algorithms are expressed naturally
as meta-algorithms parametrized on simpler algorithms.
Examples include ensemble methods which
build and combine several simpler models (e.g., decision trees), or multiclass
and multilabel classification schemes which can be used to turn a binary
classifier into a multiclass or multilabel classifier. In \sklearn, such algorithms are
implemented as \textit{meta-estimators}. They take as input an existing base
estimator and use it internally for learning and making predictions.
All meta-estimators implement the regular estimator interface.

As an example, a logistic regression classifier
uses by default a one-vs.-rest scheme
for performing multiclass classification.
A different scheme can be achieved
by a meta-estimator wrapping a logistic regression estimator:
\begin{lstlisting}
from sklearn.multiclass import OneVsOneClassifier

ovo_lr = OneVsOneClassifier(LogisticRegression(penalty="l1"))
\end{lstlisting}
For learning, the \texttt{OneVsOneClassifier} object
\textit{clones} the logistic regression estimator multiple times,
resulting in a set of $\frac{K(K-1)}{2}$ estimator objects
for $K$-way classification,
all with the same settings.
For predictions, all estimators perform a binary classification and then vote to make the final decision.
The snippet exemplifies the importance
of separating object instantiation and actual learning.

Since meta-estimators require users to construct nested objects,
the decision to implement a meta-estimator
rather than integrate the behavior it implements
into existing estimators classes
is always based on a trade-off between generality and ease of use.
Relating to the example just given,
all \sklearn classifiers are designed to do multiclass classification
and the use of the \texttt{multiclass} module
is only necessary in advanced use cases.


\subsection{Pipelines and feature unions}

A distinguishing feature of the \sklearn API is its ability to
compose new estimators from several base estimators. Composition mechanisms can
be used to combine typical machine learning workflows into a single object which
is itself an estimator, and can be employed wherever usual estimators can be used.
In particular, \sklearn's model selection routines
can be applied to composite estimators, allowing global optimization
of all parameters in a complex workflow.
Composition of estimators can be done in two
ways: either sequentially through \texttt{Pipeline} objects, or in a parallel
fashion through \texttt{FeatureUnion} objects.

\texttt{Pipeline} objects chain multiple estimators into a single one. This is
useful since a machine learning workflow typically involves a fixed sequence of
processing steps (e.g., feature extraction, dimensionality reduction, learning
and making predictions), many of which perform some kind of learning.
A sequence of $N$ such steps can be combined into a
\texttt{Pipeline} if the first $N-1$ steps are transformers; the last can be
either a predictor, a transformer or both.

Conceptually, fitting a pipeline to
a training set amounts to the following recursive procedure: i) when only one
step remains, call its \texttt{fit} method; ii) otherwise, \texttt{fit} the
first step, use it to \texttt{transform} the training set and \texttt{fit} the
rest of the pipeline with the transformed data. The pipeline exposes all the
methods the last estimator in the pipe exposes. That is, if the last estimator
is a predictor, the pipeline can itself be used as a predictor. If the last
estimator is a transformer, then the pipeline is itself a transformer.

\texttt{FeatureUnion} objects combine multiple transformers into a single one
that concatenates their outputs. A union of two transformers that
map input having $d$ features to $d'$ and $d''$ features respectively is
a transformer that maps its $d$ input features to $d' + d''$ features.
This generalizes in the obvious way to more than two transformers.
In terms of API, a \texttt{FeatureUnion} takes as input a list of transformers.
Calling \texttt{fit} on the union is the same as calling \texttt{fit}
independently on each of the transformers and then joining their outputs.

\texttt{Pipeline} and \texttt{FeatureUnion} can be
combined to create complex and nested workflows.
The following snippet illustrates how to create a complex estimator
that computes both linear PCA and kernel PCA features on \texttt{X\_train}
(through a \texttt{FeatureUnion}),
selects the 10 best features in the combination according to an ANOVA test
and feeds those to an $\ell_2$-regularized logistic regression model.
\begin{lstlisting}
from sklearn.pipeline import FeatureUnion, Pipeline
from sklearn.decomposition import PCA, KernelPCA
from sklearn.feature_selection import SelectKBest

union = FeatureUnion([("pca", PCA()),
                      ("kpca", KernelPCA(kernel="rbf"))])

Pipeline([("feat_union", union),
          ("feat_sel", SelectKBest(k=10)),
          ("log_reg", LogisticRegression(penalty="l2"))
         ]).fit(X_train, y_train).predict(X_test)
\end{lstlisting}

\subsection{Model selection}

As introduced in Section~\ref{sec:estimators}, hyper-parameters set in the
constructor of an estimator
determine the behavior of the learning algorithm
and hence the performance of the resulting model on unseen data.
The problem of \textit{model selection} is therefore to find, within
some hyper-parameter space, the best combination of hyper-parameters, with
respect to some user-specified criterion. For example, a decision
tree with too small a value for the maximal tree depth
parameter will tend to underfit, while too large a value will make it overfit.

In \sklearn, model selection is supported in two distinct meta-estimators,
\texttt{GridSearchCV} and \texttt{RandomizedSearchCV}.  They take as input an
estimator (basic or composite), whose hyper-parameters must be optimized, and a
set of hyperparameter settings to search through.
This set is represented as a mapping of parameter names
to a set of discrete choices in the case of grid search,
which exhaustively enumerates the ``grid'' (cartesian product)
of complete parameter combinations.
Randomized search is a smarter algorithm
that avoids the combinatorial explosion in grid search
by sampling a fixed number of times from its parameter distributions
(see \citealp{bergstra2012}).

Optionally, the model selection algorithms
also take a cross-validation scheme and a score function.  \sklearn provides
various such cross-validation schemes, including $k$-fold (default),
stratified $k$-fold and leave-one-out.
The score function used by default is the estimator's \texttt{score} method,
but the library provides a variety of
alternatives that the user can choose from,
including accuracy, AUC and $F_1$ score for classification,
$R^2$ score and mean squared error for regression.

For each hyper-parameter combination and each train/validation split
generated by the cross-validation scheme, \texttt{GridSearchCV}
and \texttt{RandomizedSearchCV} fit their base estimator on the training set and
evaluate its performance on the validation set.  In the end, the best performing
model on average is retained and exposed as the public attribute
\texttt{best\_estimator\_}.

The snippet below illustrates how to find
hyper-parameter settings for an SVM classifier (SVC)
that maximize $F_1$ score
through 10-fold cross-validation on the training set.
\begin{lstlisting}
from sklearn.grid_search import GridSearchCV
from sklearn.svm import SVC

param_grid = [
  {"kernel": ["linear"], "C": [1, 10, 100, 1000]},
  {"kernel": ["rbf"], "C": [1, 10, 100, 1000],
   "gamma": [0.001, 0.0001]}
]

clf = GridSearchCV(SVC(), param_grid, scoring="f1", cv=10)
clf.fit(X_train, y_train)
y_pred = clf.predict(X_test)
\end{lstlisting}
In this example, two distinct hyper-parameter grids are
considered for the linear and radial basis function (RBF) kernels;
an SVM with a linear kernel accepts a $\gamma$ parameter, but ignores it,
so using a single parameter grid would waste computing time
trying out effectively equivalent settings.
Additionally, we see that
\texttt{GridSearchCV} has a \texttt{predict} method, just like any other classifier:
it delegates the \texttt{predict}, \texttt{predict\_proba}, \texttt{transform} and
\texttt{score} methods to the best estimator
(optionally after re-fitting it on the whole training set).

\subsection{Extending scikit-learn}

To ease code reuse, simplify implementation and skip the introduction of
superfluous classes, the Python principle of \textit{duck typing} is exploited
throughout the codebase. This means that estimators are defined by interface,
not by inheritance, where the interface is entirely implicit
as far as the programming language is concerned.
Duck typing allows both for extensibility and
flexibility: as long as an estimator follows the API and conventions
outlined in Section~\ref{sec:core-api}, then it can be used in lieu of a
built-in estimator (e.g., it can be plugged into pipelines or grid search)
and external developers are not forced to inherit from any \sklearn class.

In other places of the library, in particular in the vectorization code
for unstructured input, the toolkit is also designed to be
extensible. Here, estimators provide hooks for user-defined code: objects or
functions that follow a specific API can be given as arguments at vectorizer
construction time. The library then calls into this code, communicating with it by passing objects of standard Python/NumPy types.
Again, such external user code can be kept agnostic of the \sklearn
class hierarchy.

Our rule of thumb is that user code should not be tied to \sklearn---which is a
\textit{library}, and not a \textit{framework}. This principle indeed avoids a
well-known problem with object-oriented design, which is that users wanting a
``banana'' should not get ``a gorilla holding the banana and the entire jungle''
(J.~Armstrong, cited by \citealp[p.~213]{seibel2009coders}).
That is, programs using \sklearn should not be intimately tied to it,
so that their code can be reused with other toolkits or in other contexts.

\section{Implementation}
\label{sec:implementation}

Our implementation guidelines emphasize writing efficient but readable code.
In particular, we focus on making the codebase easily maintainable and
understandable in order to favor external contributions. Whenever practicable,
algorithms implemented in \sklearn are written in Python,
using NumPy vector operations for numerical work.
This allows for the code to remain concise, readable and
efficient. For critical algorithms that cannot be easily and efficiently
expressed as NumPy operations, we rely on Cython \citep{behnel2011cython}
to achieve competitive performance and scalability. Cython is a
compiled programming language that extends Python with static typing. It
produces efficient C extension modules that are directly importable from the
Python run-time system. Examples of algorithms written in Cython include
stochastic gradient descent for linear models, some graph-based clustering
algorithms and decision trees.

To facilitate the installation and thus adoption of \sklearn,
the set of external dependencies is kept to a bare minimum:
only Python, NumPy and SciPy are required for a functioning installation.
Binary distributions of these are available for the major platforms.
Visualization functionality depends on Matplotlib \citep{hunter2007matplotlib}
and/or Graphviz \citep{gansner2000},
but neither is required to perform machine learning or prediction.
When feasible, external libraries are integrated into the codebase.
In particular, \sklearn includes modified versions of \textsf{LIBSVM} and \textsf{LIBLINEAR}
\citep{chang2011libsvm,fan2008}, both written in C{}\verb!++!
and wrapped using Cython modules.

\section{Related software}
\label{sec:comparison}

Recent years have witnessed a rising interest in machine learning and data
mining with applications in many fields. With this rise comes a host of machine
learning packages (both open source and proprietary) with which \sklearn
competes. Some of those, including Weka~\citep{hall2009weka} or
Orange~\citep{Demsar2004}, offer APIs but actually focus on the use of a graphical user interface (GUI)
which allows novices to easily apply machine learning algorithms. By
contrast, the target audience of \sklearn is capable of programming, and
therefore we focus on developing a usable and consistent API, rather than expend
effort into creating a GUI\@. In addition, while GUIs are useful tools, they
sometimes make reproducibility difficult in the case of complex workflows
(although those packages usually have developed a GUI for managing complex
tasks).

Other existing machine learning packages
such as SofiaML\footnote{\url{https://code.google.com/p/sofia-ml}}~\citep{sculley2009large}
and Vowpal~Wabbit\footnote{\url{http://hunch.net/$\sim$vw}}
are intended to be used as command-line tools
(and sometimes do not offer any type of API).
While these packages have the advantage
that their users are not tied to a particular programming language,
the users will find that they still need programming to process input/output,
and will do so in a variety of languages.
By contrast, \sklearn allows users to implement that entire workflow
in a single program, written in a single language,
and developed in a single working environment.
This also makes it easier for researchers and developers
to exchange and collaborate on software, as dependencies and setup are kept to a
minimum.

Similar benefits hold in the case of specialized languages
for numeric and statistical programming
such as \textsc{matlab} and R \citep{r}.
In comparison to these, though, Python has the distinct advantage
that it is a \textit{general purpose} language,
while NumPy and SciPy extend it with functionality
similar to that offered by \textsc{matlab} and R.
Python has strong language and standard library support for such tasks as
string/text processing, interprocess communication, networking
and many of the other auxiliary tasks that machine learning programs
(whether academic or commercial) routinely need to perform.
While support for many of these tasks is improving in languages such as
\textsc{matlab} and R, they still lag behind Python in their general purpose
applicability.
In many applications of machine learning these tasks, such as data access,
data preprocessing and reporting, can be a more significant task than applying
the actual learning algorithm.

Within the realm of Python,
a package that deserves mention is the Gensim topic modeling toolkit
\citep{rehurek2010gensim},
which exemplifies a different style of API design
geared toward scalable processing of ``big data''.
Gensim's method of dealing with large datasets is to use algorithms
that have $O(1)$ space complexity and can be updated online.
The API is designed around the Python concept of an \textit{iterable}
(supported in the language by a restricted form of co-routines called
\textit{generators}).
While the text vectorizers part of \sklearn
also use iterables to some extent,
they still produce entire sparse matrices, intended to be used for batch or
mini-batch learning. This is the case
even in the stateless, O(1) memory vectorizers
that implement the hashing trick of \citet{weinberger2009}.
This way of processing, as argued earlier in Section~\ref{sec:arrays},
reduces various forms of overhead
and allows effective use of the vectorized operations provided by NumPy and
SciPy.  We make no attempt to hide this batch-oriented processing from the user,
allowing control over the amount of memory actually dedicated
to \sklearn algorithms.

\section{Future directions}
\label{sec:future_work}
There are several directions that the \sklearn project
aims to focus on in future development.
At present, the library does not support some classical machine learning
algorithms,
including neural networks, ensemble meta-estimators for
bagging or subsampling strategies and missing value completion algorithms.
However, tasks like structured prediction or reinforcement learning are
considered out of scope for the project,
since they would require quite different data representations and APIs.

At a lower-level, parallel processing is a potential point of improvement.
Some estimators in \sklearn are already able to leverage multicore processors,
but only in a coarse-grained fashion.
At present, parallel processing is difficult to accomplish in the Python environment;
\sklearn targets the main implementation, CPython,
which cannot execute Python code on multiple CPUs simultaneously.
It follows that any parallel task decomposition must either be done
inside Cython modules,
or at a level high enough to warrant the overhead
of creating multiple OS-level processes,
and the ensuing inter-process communication.
Parallel grid search is an example of the latter approach
which has already been implemented.
Recent versions of Cython include support for the OpenMP standard
\citep{dagum1998openmp},
which is a viable candidate technology
for more fine-grained multicore support in \sklearn.

Finally, a long-term solution for model persistence is missing.
Currently, Python's \texttt{pickle} module is recommended for serialization,
but this only offers a file format,
not a way of preserving compatibility between versions.
Also, it has security problems because its deserializer
may execute arbitrary Python code,
so models from untrusted sources cannot be safely ``unpickled''.

These API issues will be addressed in the future in preparation for
the 1.0 release of \sklearn.

\section{Conclusion}
\label{sec:conclusions}

We have discussed the \sklearn API
and the way it maps machine learning concepts and tasks
onto objects and operations in the Python programming language.
We have shown how a consistent API across the package makes \sklearn
very \textbf{usable} in practice: experimenting with different learning
algorithm is as simple as substituting a new class definition.
Through composition interfaces such as Pipelines, Feature Unions,
and meta-estimators, these simple building blocks lead to an API which is
\textbf{powerful}, and can accomplish a wide variety of learning tasks
within a small amount of easy-to-read code.
Through duck-typing, the consistent API leads to a library that is
easily \textbf{extensible}, and allows user-defined estimators to be
incorporated into the \sklearn workflow without any explicit object
inheritance.

While part of the \sklearn API is necessarily Python-specific,
core concepts may be applicable to
machine learning applications and toolkits
written in other (dynamic) programming languages.
The power, and extensibility of the \sklearn API is evidenced
by the large and growing user-base, its use to solve real
problems across a wide array of fields,
as well as the appearance of third-party packages
that follow the \sklearn conventions. Examples of such packages include
\textit{astroML}\footnote{\url{http://astroml.org}}
\citep{vanderplas2012astroML}, a package providing
machine learning tools for astronomers, and
\textit{wiseRF}\footnote{\url{http://wise.io}}, a commercial random forest
implementation. The source code of
the recently-proposed sparse multiclass
algorithm of \citet{mblondel-mlj2013}, released as
part of the
\textit{lightning}\footnote{\url{https://github.com/mblondel/lightning}}
package, also follows the \sklearn conventions.
To maximize ease of use, we encourage more researchers
to follow these conventions when releasing their software.

\subsection*{Acknowledgments}

The authors and contributors acknowledge active support from INRIA\@. Past and
present sponsors of the project also include Google for funding
scholarships through its Summer of Code program,
the Python Software Foundation and Tinyclues for funding coding sprints.

Gilles~Louppe and Arnaud~Joly are research fellows of the Belgian
Fonds de la Recherche Scientifique (FNRS)
and acknowledge its financial support.

{\small
\bibliographystyle{abbrvnat}
\DeclareRobustCommand{\VAN}[3]{#3}
\bibliography{paper}

\begin{thebibliography}{21}
\providecommand{\natexlab}[1]{#1}
\providecommand{\url}[1]{\texttt{#1}}
\expandafter\ifx\csname urlstyle\endcsname\relax
  \providecommand{\doi}[1]{doi: #1}\else
  \providecommand{\doi}{doi: \begingroup \urlstyle{rm}\Url}\fi

\bibitem[Behnel et~al.(2011)Behnel, Bradshaw, Citro, Dalcin, Seljebotn, and
  Smith]{behnel2011cython}
S.~Behnel, R.~Bradshaw, C.~Citro, L.~Dalcin, D.~S. Seljebotn, and K.~Smith.
\newblock Cython: the best of both worlds.
\newblock \emph{Comp. in Sci. \& Eng.}, 13\penalty0 (2):\penalty0 31--39, 2011.

\bibitem[Bergstra and Bengio(2012)]{bergstra2012}
J.~Bergstra and J.~Bengio.
\newblock Random search for hyper-parameter optimization.
\newblock \emph{JMLR}, 13:\penalty0 281--305, 2012.

\bibitem[Blondel et~al.(2013)Blondel, Seki, and Uehara]{mblondel-mlj2013}
M.~Blondel, K.~Seki, and K.~Uehara.
\newblock Block coordinate descent algorithms for large-scale sparse multiclass
  classification.
\newblock \emph{Machine Learning}, 93\penalty0 (1):\penalty0 31--52, 2013.

\bibitem[Chang and Lin(2011)]{chang2011libsvm}
C.-C. Chang and C.-J. Lin.
\newblock {LIBSVM}: a library for support vector machines.
\newblock \emph{ACM Trans. on Intelligent Systems and Technology}, 2\penalty0
  (3):\penalty0 27, 2011.

\bibitem[Dagum and Menon(1998)]{dagum1998openmp}
L.~Dagum and R.~Menon.
\newblock {OpenMP}: an industry standard {API} for shared-memory programming.
\newblock \emph{Computational Sci. \& Eng.}, 5\penalty0 (1):\penalty0 46--55,
  1998.

\bibitem[Dem\v{s}ar et~al.(2004)Dem\v{s}ar, Zupan, Leban, and Curk]{Demsar2004}
J.~Dem\v{s}ar, B.~Zupan, G.~Leban, and T.~Curk.
\newblock Orange: From experimental machine learning to interactive data
  mining.
\newblock In \emph{Knowledge Discovery in Databases PKDD 2004}, Lecture Notes
  in Computer Science. Springer, 2004.

\bibitem[Fan et~al.(2008)Fan, Chang, Hsieh, Wang, and Lin]{fan2008}
R.-E. Fan, K.-W. Chang, C.-J. Hsieh, X.-R. Wang, and C.-J. Lin.
\newblock {LIBLINEAR}: A library for large linear classification.
\newblock \emph{JMLR}, 9:\penalty0 1871--1874, 2008.

\bibitem[Gansner and North(2000)]{gansner2000}
E.~R. Gansner and S.~C. North.
\newblock An open graph visualization system and its applications to software
  engineering.
\newblock \emph{Software---Practice and Experience}, 30\penalty0 (11):\penalty0
  1203--1233, 2000.

\bibitem[Guazzelli et~al.(2009)Guazzelli, Zeller, Lin, and
  Williams]{guazzelli2009pmml}
A.~Guazzelli, M.~Zeller, W.-C. Lin, and G.~Williams.
\newblock Pmml: An open standard for sharing models.
\newblock \emph{The R Journal}, 1\penalty0 (1):\penalty0 60--65, 2009.

\bibitem[Haenel et~al.(2013)Haenel, Gouillart, and
  Varoquaux]{varoquaux2013scipy}
V.~Haenel, E.~Gouillart, and G.~Varoquaux.
\newblock Python scientific lecture notes, 2013.
\newblock URL \url{http://scipy-lectures.github.io/}.

\bibitem[Hall et~al.(2009)Hall, Frank, Holmes, Pfahringer, Reutemann, and
  Witten]{hall2009weka}
M.~Hall, E.~Frank, G.~Holmes, B.~Pfahringer, P.~Reutemann, and I.~H. Witten.
\newblock The {WEKA} data mining software: an update.
\newblock \emph{ACM SIGKDD Explorations Newsletter}, 11\penalty0 (1):\penalty0
  10--18, 2009.

\bibitem[Hunter(2007)]{hunter2007matplotlib}
J.~D. Hunter.
\newblock Matplotlib: A 2d graphics environment.
\newblock \emph{Comp. in Sci. \& Eng.}, pages 90--95, 2007.

\bibitem[Pedregosa et~al.(2011)Pedregosa, Varoquaux, Gramfort, Michel, Thirion,
  Grisel, Blondel, Prettenhofer, Weiss, Dubourg, Vanderplas, Passos,
  Cournapeau, Brucher, Perrot, and Duchesnay]{pedregosa2011}
F.~Pedregosa, G.~Varoquaux, A.~Gramfort, V.~Michel, B.~Thirion, O.~Grisel,
  M.~Blondel, P.~Prettenhofer, R.~Weiss, V.~Dubourg, J.~Vanderplas, A.~Passos,
  D.~Cournapeau, M.~Brucher, M.~Perrot, and E.~Duchesnay.
\newblock Scikit-learn: Machine learning in {P}ython.
\newblock \emph{JMLR}, 12:\penalty0 2825--2830, 2011.

\bibitem[Perez and Granger(2007)]{perez2007ipython}
F.~Perez and B.~E. Granger.
\newblock {IPython}: a system for interactive scientific computing.
\newblock \emph{Comp. in Sci. \& Eng.}, 9\penalty0 (3):\penalty0 21--29, 2007.

\bibitem[{R Core Team}(2013)]{r}
{R Core Team}.
\newblock \emph{R: A Language and Environment for Statistical Computing}.
\newblock R Foundation, Vienna, Austria, 2013.
\newblock URL \url{http://www.R-project.org}.

\bibitem[{\v R}eh{\r u}{\v r}ek and Sojka(2010)]{rehurek2010gensim}
R.~{\v R}eh{\r u}{\v r}ek and P.~Sojka.
\newblock Software framework for topic modelling with large corpora.
\newblock In \emph{Proc. LREC Workshop on New Challenges for NLP Frameworks},
  pages 46--50, 2010.

\bibitem[Sculley(2009)]{sculley2009large}
D.~Sculley.
\newblock Large scale learning to rank.
\newblock In \emph{NIPS Workshop on Advances in Ranking}, pages 1--6, 2009.

\bibitem[Seibel(2009)]{seibel2009coders}
P.~Seibel.
\newblock \emph{Coders at Work: Reflections on the Craft of Programming}.
\newblock Apress, 2009.

\bibitem[{Vanderplas} et~al.(2012){Vanderplas}, {Connolly}, {Ivezi{\'c}}, and
  {Gray}]{vanderplas2012astroML}
J.~{Vanderplas}, A.~{Connolly}, {\v Z}.~{Ivezi{\'c}}, and A.~{Gray}.
\newblock Introduction to {astroML}: Machine learning for astrophysics.
\newblock In \emph{Conf. on Intelligent Data Understanding (CIDU)}, pages
  47--54, 2012.

\bibitem[{\VAN{Walt}{Van der}{van der}}~Walt et~al.(2011){\VAN{Walt}{Van
  der}{van der}}~Walt, Colbert, and Varoquaux]{vanderwalt2011}
S.~{\VAN{Walt}{Van der}{van der}}~Walt, S.~C. Colbert, and G.~Varoquaux.
\newblock The {NumPy} array: a structure for efficient numerical computation.
\newblock \emph{Comp. in Sci. \& Eng.}, 13\penalty0 (2):\penalty0 22--30, 2011.

\bibitem[Weinberger et~al.(2009)Weinberger, Dasgupta, Langford, Smola, and
  Attenberg]{weinberger2009}
K.~Weinberger, A.~Dasgupta, J.~Langford, A.~Smola, and J.~Attenberg.
\newblock Feature hashing for large scale multitask learning.
\newblock In \emph{Proc. ICML}, 2009.

\end{thebibliography}
}

\end{document}